## Title

Koopman global linearization of contact dynamics for robot locomotion and manipulation enables elaborate control

## Authors

C. O'Neill,[1]* J. Terrones,[1] H. H. Asada[1]

## Affiliations

[1]Department of Mechanical Engineering, MIT, Cambridge, 02139, MA
These authors contributed equally: C. O'Neill, J. Terrones
*Corresponding author. Email: croneill@mit.edu

## Abstract

Controlling robots that dynamically engage in contact with their environment is a pressing challenge. Whether a legged robot making-and-breaking contact with a floor, or a manipulator grasping objects, contact is everywhere. Unfortunately, the switching of dynamics at contact boundaries makes control difficult. Predictive controllers face non-convex optimization problems when contact is involved. Here, we overcome this difficulty by applying Koopman operators to subsume the segmented dynamics due to contact changes into a unified, globally-linear model in an embedding space. We show that viscoelastic contact at robot-environment interactions underpins the use of Koopman operators without approximation to control inputs. This methodology enables the convex Model Predictive Control of a legged robot, and the real-time control of a manipulator engaged in dynamic pushing. In this work, we show that our method allows robots to discover elaborate control strategies in real-time over time horizons with multiple contact changes, and the method is applicable to broad fields beyond robotics.

## Introduction

Robots perform tasks through interactions with their environment. Particularly challenging are those tasks in which the robot makes and breaks contact. Such cases are commonplace throughout robotics: legged robots repeat a cycle of ground contact/non-contact for each leg, and manipulators make and break contact with an object. Task planning and control strategy synthesis for these systems is complex; significant changes to their dynamic behaviors occur as contact conditions vary, and the governing equations must be switched accordingly. While this switching lends itself to a segmented representation of the dynamics, global behaviors are difficult to elucidate from a group of segmented equations, and control strategies for navigating through switching events are even harder to synthesize. Analytic tools are limited for predicting and controlling such switched nonlinear systems[1–6] and traditional control techniques are oftentimes inapplicable to systems undergoing contact[7].

Model Predictive Control (MPC) has been considered in both locomotion and manipulation for controlling contact-rich systems. Prior works have demonstrated successful results by applying MPC to systems with contact under restrictive conditions and assumptions. Representing contact dynamics within an MPC optimization leads to mixed-integer or linear-complementarity problems, which are nonlinear and non-convex. Such computational methods are challenging to implement, especially at real-time rates on hardware, due to coupling between contact dynamics and control inputs[8]. Compliant models have been found to aid in optimization due to gradients, but can still be challenging to run in real-time[9]. Consequently, past work that has applied MPC for real-time control has leaned on certain simplifications, such as:

a) Simplified Dynamics

   i. Quasi-dynamic or quasi-static models[1,2,10–12]

   ii. Models of reduced dimensionality and complexity[13–18]

b) Linearization at a reference point or along a reference trajectory[19]

c) Pre-defined contact mode sequences[11,12,18,20–23]

Each of these assumptions limits the controller's behavior. The use of quasi-dynamic models, (a-i), prevents MPC from exploring dynamic behaviors. With assumption (a-ii), the use of a reduced order model within MPC requires a low-level controller to translate the reduced order trajectory into whole-body commands[14,18,24]. Furthermore, the simplifications in the dynamics and the lack of constraints inherent to reduced order models can yield infeasible or suboptimal motions[18]. Assumption (b), the use of local linearization, inhibits an MPC controller from exploring control broadly in the task space. Such linearization fails to approximate dynamics across contact mode boundaries, making it difficult to find a policy that takes advantage of contact-rich trajectories. Pre-defining a contact sequence in advance, (c), requires an offline motion planner to provide the necessary contact modes. Even when a contact mode sequence is not explicitly pre-defined, bias terms may nevertheless be used to guide contact actions[25] Relying upon contact modes that are defined prior to runtime can also lead to poor control performance when facing unexpected changes in contact mode.

Ultimately, these assumptions stem from systems with contact dynamics, whether involved in locomotion or manipulation tasks, lacking a global, unified model that is amenable for robot planning and control. Here, we propose a unified modeling method based on Koopman operator theory, so that the robot can explore broadly effective control strategies that may involve multiple switching events.

Koopman operators[26,27] have the potential to provide us with globally linear dynamic models in an embedding space that can subsume a class of segmented dynamic equations[28]. The globally linear models facilitate the use of the wealth of linear systems theory and tools. When applied to MPC, for example, computation becomes simple convex-optimization with no local minima, enabling real-time computation[29]. If applied properly, the Koopman model enables linear MPC to predict trajectories involving a series of contact switches over a time horizon. The controller can then obtain an optimal sequence of control inputs that may cross the boundaries of segmented dynamic equations in the original state space. With this method, we can remove the need for Linear Complementarity Constraints or Mixed Integer Programming while also removing the simplifying assumptions previously required for real-time MPC.

To make this Koopman-based modeling possible, two fundamental challenges must be overcome. One is that the original Koopman operator theory is applicable only to autonomous systems with no control input. The second fundamental challenge is that the existence of a Koopman operator cannot be guaranteed for dynamic systems with discontinuities.

To resolve the first challenge, at least three methods have been proposed. They are a) least square estimation of both system matrix and control matrix for given data of input and state variables, e.g. Dynamic Mode Decomposition with Control (DMDc)[30], b) the method to treat control inputs as part of the independent state variables and apply the standard Koopman Operator to the augmented state variables[31], and c) bilinear approximation of control terms[32–34]. In most applications, the control matrix varies depending on the state. Approximating the control terms to a constant matrix results in significant errors. The bilinear approximation produces a more accurate approximation[35,36], but the resulting model is no longer linear. Dynamic analysis and control of bilinear systems are more restrictive and complex, compared to the completely linear Koopman model.

If we deal with general nonlinear systems, Koopman operator theory cannot be applied to non-autonomous systems without approximation. However, there is a class of nonlinear control systems that possess a particular structure for which Koopman operators can be obtained without approximation to the control matrix. The method referred to as Control-Coherent Koopman (CCK) modeling, originally developed for systems with powertrain compliance[37], allows us to construct a discrete-time, Koopman operator with a constant control matrix that is valid globally for such systems. The current work extends and improves the original CCK formulation[35], and addresses how contact-rich locomotion and manipulation processes can be reduced to Koopman-compatible physical models to which CCK modeling can be applied.

The second fundamental challenge in applying Koopman operator theory is discontinuity at switching events. Many legged robots are modeled as hybrid systems where ground collisions are treated as discrete impulses, leading to instantaneous contact dynamics and discontinuous changes in momenta. In general, these discontinuities in the governing equations prohibit the application of Koopman operator theory: they violate one of the key assumptions of the Koopman operator[27].

The discontinuity of momenta arising from hybrid modeling is, in part, a consequence of modeling decisions. In real physical systems, which are causal, momentum is continuous although the time rate of change is large. At a microscopic level, the floor impact is not discontinuous, and a compliant contact yields more realistic modeling results[38]. Contact forces cannot always be modeled as motion constraints, especially in cases where deformation matters, such as when there are indeterminate contact forces or simultaneous collisions[39]. The human gait, too, has some compliance at the feet and the legs when compared to rigid body hybrid models[40]. Hybrid models ignore the viscoelastic properties at the floor-leg contact, resulting in discontinuity in momenta. This can be overcome by considering causal physical models with contact compliance that are proven to be realistic and practical.

In this work, we establish an alternative approach to the modeling, prediction, and control of complex robot systems that make-and-break contact. We will show that Control-Coherent Koopman (CCK) and causal contact modeling with compliance allows us to obtain a globally linear, unified model that subsumes a group of segmented dynamic models and provides MPC-amenable representations of otherwise highly complex, switched nonlinear systems. The resultant Koopman-MPC controller can predict trajectories that include a series of contact mode switches over a time horizon and find an optimal control sequence in real time. This optimization includes strategic decisions in the selection of effective contact sequences, coordination of multiple axes, and types of contact actions. The unified, globally-linear modeling enables the robot to discover effective strategies in real-time. We will address this for both locomotion and manipulation: two major fields of robotics. The method is applicable to broad fields beyond robotics where governing dynamics are segmented.

## Results

### Overview

In our proposed method we analyze and represent global behaviors of segmented dynamics of robotic systems in the light of Koopman operator theory, construct Koopman-compatible models, and apply linear MPC. This is outlined in Fig. 1. The focus is dynamic modeling and control of switched robot systems, where the governing equations of motion are segmented in the state space. These are analyzed and made into a causal physical model to which Koopman operator theory is applicable. Combined with the Control-Coherent Koopman framework, Koopman-compatible physical models are constructed. The resultant model is a linear and time-invariant system, which enables linear, convex MPC. Finally, CCK-MPC is applied to systems undergoing contact; exemplary cases in legged locomotion and manipulation are studied.

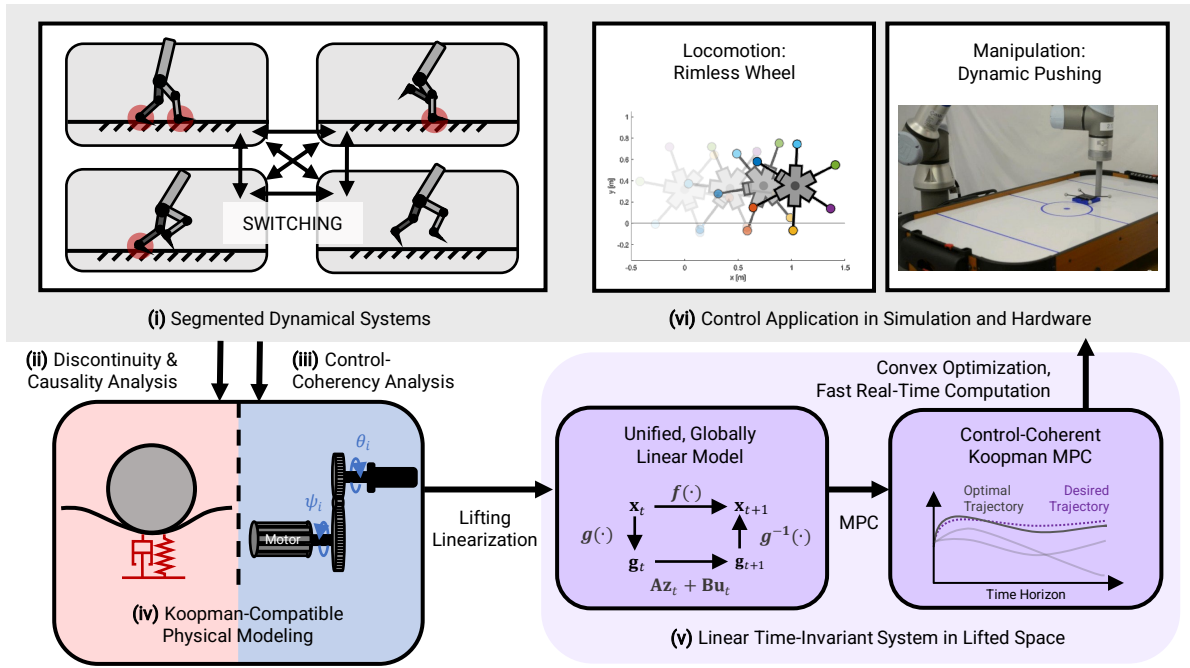

**Figure 1. General presented framework**
Overview of the proposed CCK-MPC method: (**i**) Systems that make-and-break contact have governing equations of motion that are segmented in the state space. (**ii**) We analyze the switching dynamics and derive a causal physical model to which Koopman operator theory is applicable. (**iii**) Furthermore, we analyze under which conditions the Koopman operator theory can be extended to non-autonomous systems with control. (**iv**) Based on these analyses, Koopman-compatible physical models are constructed, which are ready for lifting linearization. (**v**) The resultant model is a linear, time-invariant system, which makes MPC control straightforward and effective. (**vi**) The proposed CCK-MPC approach is applied to two exemplary case studies in legged locomotion and manipulation.

In the following, we address the applicability of Koopman operator theory and obtain the conditions for Koopman-compatible modeling. We will demonstrate how unified models subsuming segmented dynamics can help MPC controllers predict the state evolution beyond contact boundaries. Through numerical simulations and hardware experiments, we will show how the CCK-MPC controllers can find effective control strategies in real-time with minimum guidance for challenging locomotion and manipulation goals.

## Control-Coherent Koopman Modeling

The background mathematics of the Koopman operator theory is explained in a simple, intuitive manner in Supplementary Discussion 1.1.

Consider a discrete-time, nonlinear dynamical system with state $\mathbf{x}_t \in X \subset \mathbb{R}^n$ and control $\mathbf{u}_t \in U \subset \mathbb{R}^r$ given by

$$\mathbf{x}_{t+1} = f(\mathbf{x}_t, \mathbf{u}_t) \tag{1}$$

where $f$ is a continuously differentiable function, $f: X \times U \to X$ defined in compact subsets $X$ and $U$. This is a non-autonomous system with control to which the original Koopman operator theory does not apply[27]. It is known that Koopman operators cannot be constructed without approximating the control input terms for general nonlinear non-autonomous systems[33]. However, for a class of non-autonomous systems to be addressed in this article, Koopman operators that globally linearize the control system exist and are expressed in the following form as a Linear Time-Invariant (LTI) system.

$$\mathbf{z}_{t+1} = \mathbf{A}\mathbf{z}_t + \mathbf{B}\mathbf{u}_t \tag{2}$$

where $\mathbf{z}_t$ is the lifted state consisting of an infinite number of observables $g_i(\mathbf{x})_{i=1}^{\infty}$ that span a Hilbert space $\mathcal{H}$, and $\mathbf{A}$ and $\mathbf{B}$ are, respectively, constant system matrix and control matrix of infinite dimensions. This LTI model of non-autonomous systems is referred to as Control-Coherent Koopman (CCK) model[37].

The key to the CCK formulation is to incorporate actuator dynamics into the state equation given in Equation (1). The state vector $\mathbf{x}_t$ is divided into state variables associated with the actuator dynamics, $\mathbf{p}_t \in P \subset \mathbb{R}^m$, and the rest of the states, mainly associated with the load or plant dynamics, $\mathbf{q}_t \in X_q \subset \mathbb{R}^{n-m}$.

$$\mathbf{x}_t = \begin{bmatrix} \mathbf{p}_t \\ \mathbf{q}_t \end{bmatrix} \tag{3}$$

The class of non-autonomous systems that we consider for constructing a Control-Coherent Koopman model have the following structure.

$$\mathbf{p}_{t+1} = f_p(\mathbf{p}_t, \mathbf{q}_t) + \mathbf{B}_p \mathbf{u}_t \tag{4}$$

$$\mathbf{q}_{t+1} = f_q(\mathbf{p}_t, \mathbf{q}_t) \tag{5}$$

where $f_q: P \times X_q \mapsto X_q \subset \mathbb{R}^{n-m}$ and $f_p: P \times X_q \mapsto P$ are continuously differentiable. In this model,

- The control input appears linearly in the actuator dynamics (Modeling Property 1), and

- The actuators and the plant have independent state variables (Modeling Property 2).

If some state variables in $\mathbf{q}_t$ are algebraic functions of $\mathbf{p}_t$, Equation (5) includes the input $\mathbf{u}_t$ directly and the input appears nonlinearly. In consequence, the system is Koopman-incompatible. Property 1 can be satisfied in many actuators used in robotics and other electromechanical systems. DC motors, for example, meet this requirement because the motor rotor has a constant inertia and the motor torque appears linearly in the equations of motion of the actuator, shown in Equation (4).

Associated with the non-autonomous system, Equations (4) and (5), denoted $\mathcal{S}_\mathcal{N}$, we consider the following autonomous system $\mathcal{S}_\mathcal{A}$.

$$\begin{aligned}\bar{\mathbf{p}}_{t+1} &= f_p(\mathbf{p}_t, \mathbf{q}_t) \\ \mathbf{q}_{t+1} &= f_q(\mathbf{p}_t, \mathbf{q}_t)\end{aligned} \quad (6)$$

where $\bar{\mathbf{p}}_{t+1}$ represents the actuator state that transitions without input $\mathbf{u}_t$. This is a standard autonomous system for which a Koopman operator $\mathbf{A}$ can be constructed using observables $g_i(\mathbf{p}, \mathbf{q})_{i=1}^\infty$, whose compositions with functions $f_p(\mathbf{p}, \mathbf{q})$ and $f_q(\mathbf{p}, \mathbf{q})$ are in a Hilbert space. See Supplementary Discussion 1.1.

$$\bar{\mathbf{z}}_{t+1} = \mathbf{A}\mathbf{z}_t \quad (7)$$

In the lifted space $\mathbf{z}_t$, the dynamics evolve linearly with time. To construct a CCK model, the following two conditions are required for the above Koopman model:

- The Koopman operator of the autonomous system in Equation (7) is valid for all $\mathbf{q}_t \in X_q$ and all $\mathbf{p}_t \in P$, including all the actuator states that can be driven by an arbitrary input $\mathbf{u}_t$ in $U$, according to Equation (4), (Condition a).

- The lifted state $\mathbf{z}$ includes the actuator state $\mathbf{p}$, (Condition b).

Now, how can the autonomous system's Koopman operator $\mathbf{A}$ be combined with the linear control term in Equation (4) to obtain a LTI system in the form of Equation (2)? Condition b requires that the actuator state variables are part of the observables, which facilitates the incorporation of the linear control terms into the Koopman model. Without loss of generality, the actuator state $\mathbf{p}_t$ is placed in the first $m$ components of the lifted state variable $\mathbf{z}_t$ since $\mathbf{p}_t$ is part of $\mathbf{z}_t$:

$$\mathbf{z}_t = \begin{bmatrix} \mathbf{p}_t \\ \mathbf{g}_t \end{bmatrix} \quad (8)$$

where $\mathbf{g}_t = [g_{m+1}(\mathbf{p}_t, \mathbf{q}_t), g_{m+2}(\mathbf{p}_t, \mathbf{q}_t), \cdots]^T$ is called embeddings. Partitioning the $\mathbf{A}$ matrix into 4 blocks associated with $\mathbf{p}_t$ and $\mathbf{g}_t$, Equation (7) can be written as

$$\begin{bmatrix} \bar{\mathbf{p}}_{t+1} \\ \bar{\mathbf{g}}_{t+1} \end{bmatrix} = \begin{bmatrix} \mathbf{A}_{pp} & \mathbf{A}_{pg} \\ \mathbf{A}_{gp} & \mathbf{A}_{gg} \end{bmatrix} \begin{bmatrix} \mathbf{p}_t \\ \mathbf{g}_t \end{bmatrix} \quad (9)$$

where $\bar{\mathbf{g}}_{t+1} = \bar{\mathbf{g}}_{t+1}(\bar{\mathbf{p}}_{t+1}, \mathbf{q}_{t+1})$. Note that the nonlinear term of the actuator dynamics $f_p(\mathbf{p}_t, \mathbf{q}_t)$ corresponds to the first-row block, $\mathbf{A}_{pp}\mathbf{p}_t + \mathbf{A}_{pg}\mathbf{g}_t$, in the lifted space. Adding the linear control term $\mathbf{B}_p\mathbf{u}_t$ recovers the actuator state equation (4).

$$\mathbf{p}_{t+1} = \bar{\mathbf{p}}_{t+1} + \mathbf{B}_p\mathbf{u}_t = f_p(\mathbf{p}_t, \mathbf{q}_t) + \mathbf{B}_p\mathbf{u}_t = \mathbf{A}_{pp}\mathbf{p}_t + \mathbf{A}_{pg}\mathbf{g}_t + \mathbf{B}_p\mathbf{u}_t \tag{10}$$

The second-row block in Equation (9) corresponds to the plant state transition $f_q(\mathbf{p}_t, \mathbf{q}_t) = \mathbf{A}_{pp}\mathbf{p}_t + \mathbf{A}_{pg}\mathbf{g}_t$. Note that $\bar{\mathbf{g}}_{t+1}(\bar{\mathbf{p}}_{t+1}, \mathbf{q}_{t+1})$ depends on $\bar{\mathbf{p}}_{t+1}$ and is not a function of $\mathbf{p}_{t+1}$. To obtain the state evolution of the non-autonomous system, $\bar{\mathbf{g}}_{t+1}(\bar{\mathbf{p}}_{t+1}, \mathbf{q}_{t+1})$ must be converted to $\mathbf{g}_{t+1}(\mathbf{p}_{t+1}, \mathbf{q}_{t+1})$, corresponding to $\mathbf{p}_{t+1}$. It can be shown that this conversion, called embeddings compensation, can be made by adding a linear control term $\mathbf{B}_g\mathbf{u}_t$ with a constant gain $\mathbf{B}_g$. By combining the time evolutions of both $\mathbf{p}_t$ and $\mathbf{g}_t$, the complete state evolution of $\mathcal{S}_\mathcal{N}$ in the lifted space is given by

$$\begin{bmatrix} \mathbf{p}_{t+1} \\ \mathbf{g}_{t+1}(\mathbf{p}_{t+1}, \mathbf{q}_{t+1}) \end{bmatrix} = \begin{bmatrix} \mathbf{A}_{pp} & \mathbf{A}_{pg} \\ \mathbf{A}_{gp} & \mathbf{A}_{gg} \end{bmatrix} \begin{bmatrix} \mathbf{p}_t \\ \mathbf{g}_t \end{bmatrix} + \begin{bmatrix} \mathbf{B}_p \\ \mathbf{B}_g \end{bmatrix} \tag{11}$$

where $\mathbf{B}_g \in \mathbb{R}^{\infty \times r}$. See the Method section for the derivation of $\mathbf{B}_g$. This CCK model is LTI, which is amenable for control design.

The objective of the current work is to apply this CCK formulation to switched systems where governing equations are segmented. The dynamic ranges of $\mathbf{p}_t \in P$ and $\mathbf{q}_t \in X_q$ are segmented to $n_s$ regions $P_1 \cdots P_{n_s}$ and $X_{q1} \cdots X_{qn_s}$, respectively, and the governing equation is expressed as a collection of diverse, regional dynamics.

$$f(\mathbf{x}_t, \mathbf{u}_t) = \begin{pmatrix} f_{pi}(\mathbf{p}_t, \mathbf{q}_t) + \mathbf{B}_p\mathbf{u}_t \\ f_{qi}(\mathbf{p}_t, \mathbf{q}_t) \end{pmatrix}, \quad \mathbf{p}_t \in P_i, \ \mathbf{q}_t \in X_{qi}, \quad i = 1, \cdots, n_s \tag{12}$$

It is assumed that transitions across the borders of regions are continuous and that the Koopman operator exists for the associated autonomous system $\mathcal{S}_\mathcal{A}$. This Control-Coherent Koopman modeling is a method for aligning a nonlinear, non-autonomous dynamical system with the original Koopman operator theory. The two conditions (Conditions a and b) and the two modeling properties (Properties 1 and 2) must be satisfied to apply the CCK modeling framework. The CCK formulation is general and is applicable to all systems, including the segmented dynamical systems, satisfying the conditions. Those CCK-compatible systems include locomotion and manipulation robots of special interest, an actuated rimless wheel and a dynamic pushing robot, where the CCK requirements can be satisfied by forming causal, continuous dynamic models.

This CCK formulation is particularly useful and powerful for MPC control of segmented dynamical systems. Unlike standard local linearization, which is taken at a nominal point in a segmented region, the CCK formulation provides MPC with a globally linear model. Although the nominal point of a local linearization is close to a border of the region, the traditional linearized model contains no information about the adjacent region beyond the border. The CCK globally linear model can go beyond contact mode borders and find better trajectories. Before addressing the challenging locomotion and manipulation problems, the Koopman modeling and optimal control method is applied to a cart-pole system that makes and breaks contact with a pair of walls (Supplementary Discussion 1.3 and Supplementary Movie 7).

## Rimless Wheel Gait Dynamics

Legged robots exhibit complex and nonlinear dynamics due to the switching of governing equations as legs make-and-break contact with the ground. To construct a Koopman operator, we consider an actuated rimless wheel on a viscoelastic floor. As shown in Fig. 2a, each spoke (leg) of the rimless wheel makes-and-breaks contact in a compliant manner, which is continuous in state transitions. When no spokes are in contact, the wheel's hub behaves like a projectile body. As one spoke is in contact with the floor, it forms an open kinematic chain, and when two spokes are in contact with the floor it forms a closed-loop chain. At least these three sets of dynamic equations are switched as the rimless wheel rolls. Once a Koopman operator is constructed, these three dynamic equations are subsumed and incorporated into a single linear equation in an embedding space. There is no need to explicitly detect the contact state and distinguish which governing equation to apply.

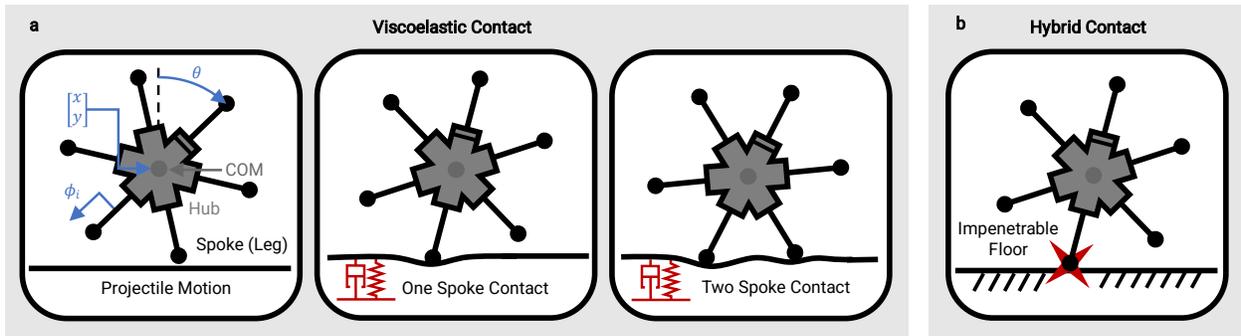

**Fig. 2 Contact modes and generalized coordinate definition of the actuated rimless wheel.**
(**a**) shows the actuated rimless wheel in different contact configurations: projectile motion, one spoke in contact with the viscoelastic floor, and two spokes in contact. The dark gray circle in the center of the hub denotes the center of mass (COM) location, given by the coordinates $r_0 = [x, y]^T$. The model's generalized coordinates also include the hub rotation angle $\theta$ and the actuator displacements for each spoke, denoted by $\phi_i$. (**b**) illustrates the causality issue from a hybrid contact model. With the impenetrable floor constraint, independent state variables simultaneously define the position of the rimless wheel, resulting in a causal conflict.

The compliance of spoke-floor interactions is essential. If it were rigid, the wheel's dynamics become discontinuous, and the existence of Koopman operators becomes questionable. The floor compliance also plays a pivotal role in satisfying the CCK formulation conditions. Let $x, y, \theta$ be the position and orientation of the hub of the rimless wheel, as shown in Fig. 2a. Each spoke is actuated with a separate actuator determining the spoke length indicated by $\phi_i$. When the spoke is not in contact with the floor, $\phi_i$ and $x, y, \theta$ are independent. However, under a rigid contact model, the hub position and some spoke lengths become kinematically dependent once a spoke contacts the ground (Fig. 2b). There is a causal conflict when multiple states dictate their position independently[41]. Introducing floor compliance resolves this conflict; both variables become independent to each other, thereby satisfying the second condition for CCK modeling, independence of actuator states (Modeling Property 2). The first condition, the linearity in input, is satisfied by the many actuators used for driving the spokes. In the current work, DC motors with lead screws are considered for modeling the actuator dynamics in Equation (4), where control inputs, i.e. armature currents, appear linearly in the actuator subsystems (Modeling Property 1).

With this modeling framework, a Koopman-compatible physical model is constructed for the rimless wheel. A CCK model in the form of Equations (8) and (9) is obtained in the following steps. First, i) the matrix $\mathbf{B}_p$ is constructed from the actuator dynamics, ii) the non-autonomous system of Equations (4) and (5) are simulated to create data, iii) using Radial Basis Functions (RBFs), the state variables are lifted, iv) the $\mathbf{A}$ matrix of $\mathcal{S}_\mathcal{A}$ is determined from simulation data, and v) the matrix $\mathbf{B}_g$ is constructed from the $\mathbf{A}$ matrix.

Note that the data used for determining $\mathbf{A}$ contains data of the three segmented equations: no-contact, one-contact, and two-contact dynamics. The three segmented equations of motion are subsumed into the linear state equation in the embedding space, Equation (2). MPC is implemented for this linear time-invariant CCK model. The CCK-MPC controller predicts the state transition over multiple switching events and implicitly determines an optimal sequence of contact mode changes. When a reference trajectory of the hub is given, CCK-MPC finds a way of coordinating the spoke movements to continuously roll in the desired direction. Furthermore, the Rimless Wheel is able to start rolling from an unfavorable initial condition with no angular momentum and both spokes on the ground. CCK-MPC discovers that by rocking back and forth, it can generate enough momentum to escape the stance posture (Supplementary Movie 1). Although the reference trajectory does not provide CCK-MPC with the rocking trajectory, the controller discovers and enacts the rocking strategy for diverse, unfavorable initial conditions. An MPC that linearizes dynamics around a reference trajectory would likely not discover the same strategy due to the unexpected contact modes.

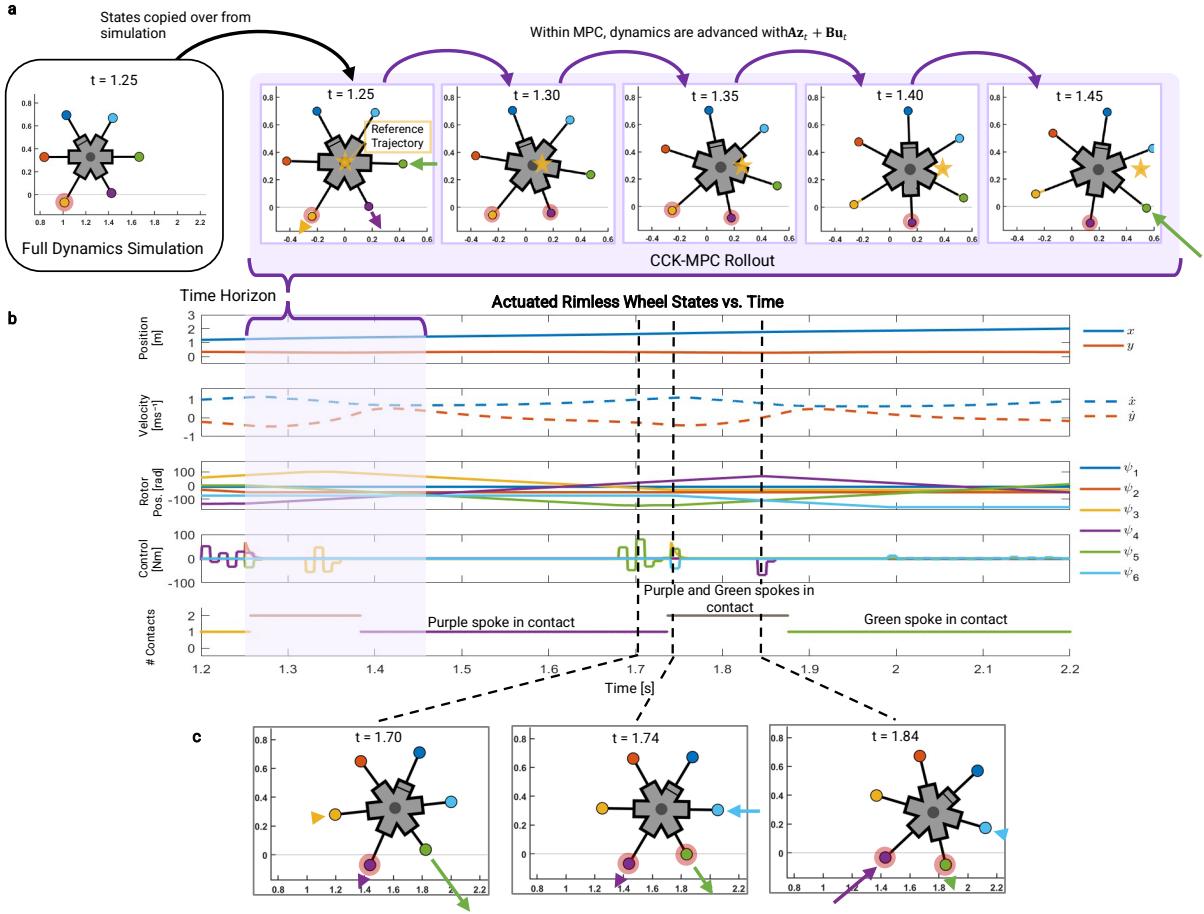

**Fig. 3 CCK-MPC enables forward rolling of the Actuated Rimless Wheel**
(**a**) shows the future states predicted using the linear Koopman model with optimized control inputs. Over the 0.2 second time horizon starting from t = 1.25 sec, CCM-MPC predicts multiple contact mode changes. Each spoke is marked with a different color, spokes in contact with the ground are marked with red circles, and control commands are represented by colored arrows. The length of each arrow reflects the magnitude of the control input. The gold star indicates the reference trajectory x position at the given time step and the dark gray circle within the hub denotes the wheel's COM. (**b**) shows the simulation states and control commands corresponding to a successful rolling trajectory. The panel also shows the number of spokes in contact with the ground. In this simulation, $\Delta_k = 0.01$ seconds and N = 20. (**c**) shows snapshots illustrating the strategy employed by the CCK-MPC to achieve continuous forward rolling, namely the aggressive extension of a spoke prior to contact to counteract expected ground reaction forces (green spoke at t = 1.70), and the early retraction of forward spokes (light blue spoke at t = 1.74 and 1.80).

Figure 3 shows the details about the rollout of states from the MPC optimization. A key strategy discovered by CCK-MPC is to proactively retract the front-most spoke. In the first panel of the CCK-MPC rollout (Fig. 3a), the green spoke is commanded to retract well before an expected collision. CCK-MPC also extends the back spoke when the spoke is positioned behind the wheel's Center of Mass (COM), injecting energy into the system without impeding the rolling motion. See the purple spoke at *t* = 1.70 sec and *t* = 1.74 sec in Fig. 3c. This combination of a lengthened back spoke and a shortened front spoke creates a ramp like effect, causing the wheel to tip forward and roll. See Supplementary Movie 2 for animations of this trajectory.

The benefits of a controller that can reason over future states are only realized when the time horizon of the controller is sufficiently long. In Fig. 3a, the CCK-MPC rollout shows two predicted contact mode changes over the time horizon of 0.2 seconds, accurately matching the nonlinear simulation, which transitions from single to double contact and back to single contact during the prediction period. When the time horizon is too small, CCK-MPC performs fewer preemptive actions and the wheel cannot continuously roll (Supplementary Movie 2).

It is important that the model be *globally* valid. Point-wise linearization methods fail because they are valid only in the vicinity of the reference points. They are too myopic to anticipate contact-mode transitions and are unable to predict the state beyond the contact transition. Figure 4a compares the performance of CCK-MPC against linear-MPC using local linearization and DMDc, two approaches that yield reasonable local models but inaccurate global predictions. Each model was evaluated with ten trials initialized along the reference trajectory and using the same MPC formulation. Only CCK-MPC could consistently achieve forward rolling. The local linearization model, unable to plan beyond its current contact mode, did not preemptively retract its forwards spokes and could not maintain its momentum (Supplementary Movie 3).

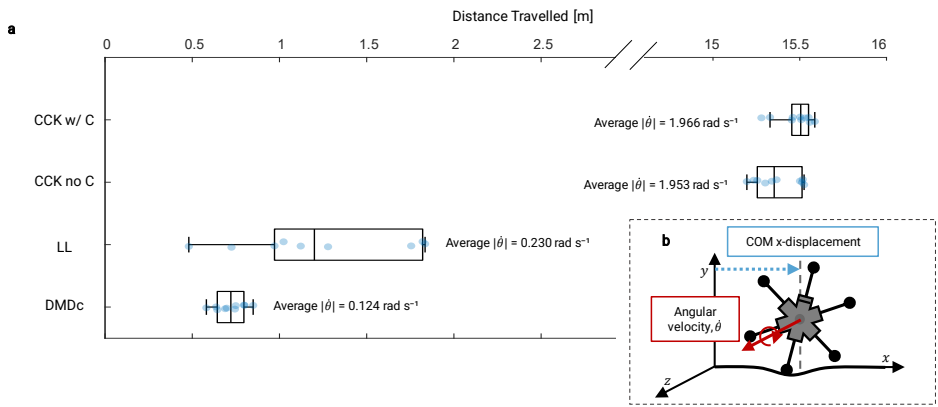

**Fig. 4 Comparison of various linearization models in L-MPC**
Shows box plots of the maximum x-displacement over the 20-second simulation, illustrating each model's ability to roll forward. Aggregated results from ten simulations for controllers based on CCK with an embedding compensation term (CCK w/ C) and without the embedding compensation term (CCK no C), as well as local linearization (LL) and DMDc, are presented. The linear MPC formulation was identical across all the four models (Supplementary Table 2). Only the CCK models were able to consistently, continuously roll, and both types achieved an average displacement greater than 15 m. The local linearization model was hindered by its inability to see beyond contact mode boundaries, and DMDc, which obtains its linear dynamics from data, did not recover the same values as the B matrix generated analytically. DMDc's erroneous B matrix impacted its prediction capabilities and worsened its control performance. The box plot summarizes data using the 25th and 75th percentiles (box), the median (line inside the box), and whiskers that extend to the furthest data points within 1.5 times the interquartile range. The blue circles denote the maximum x-displacement values for individual trials. The average absolute angular velocity is displayed for each model type. A simulation that gets stuck in a stance position may not travel far, but if it is attempting to move by rocking back in forth, it will have a higher angular velocity. **b**) defines the COM x-displacement and angular velocity metrics.

Finally, we also compare the control performance of the same CCK model with and without using the compensation term (denoted "CCK w/ C" and "CCK no C", respectively). For these ten trials, MPC with the compensation term used 5% less control effort relative to MPC without the compensation term. CCK w/ C had an overall better performance, both in terms of distance travelled and control usage (see Methods and Supplementary Methods 2.1).

## Dynamic Pushing and Shoving

The proposed CCK-MPC controller can make non-prehensile manipulation, such as pushing, truly dynamic and dexterous. Not only quasi-dynamic pushing but also shoving and dribbling of objects can be performed, where the object makes-and-breaks contact repeatedly. Figures 5a and 6a show the setup of dynamic pushing and shoving experiments with two different objects being interacted with.

The system consists of a sliding block and a controllable robot finger, called the slider and pusher, respectively. The slider can only be acted on via contact with the pusher, and experiences friction with the table during motion. The slider has three generalized coordinates, $x, y, \theta$, viewed from a global reference frame, while the pusher's position is represented with its coordinates, $x_p, y_p$. See Fig.5b. When the slider is not in contact with the pusher, the slider and the pusher possess independent generalized coordinates, as well as independent state variables. However, when contact occurs the original set of generalized coordinates are no longer independent. This results in a causal conflict, as the state of both the pusher and slider separately dictate their positions. To resolve this causal conflict, we assume some viscoelastic element between the two. Then the independent and complete set of generalized coordinates become $x, y, \theta$ and $x_p, y_p$.

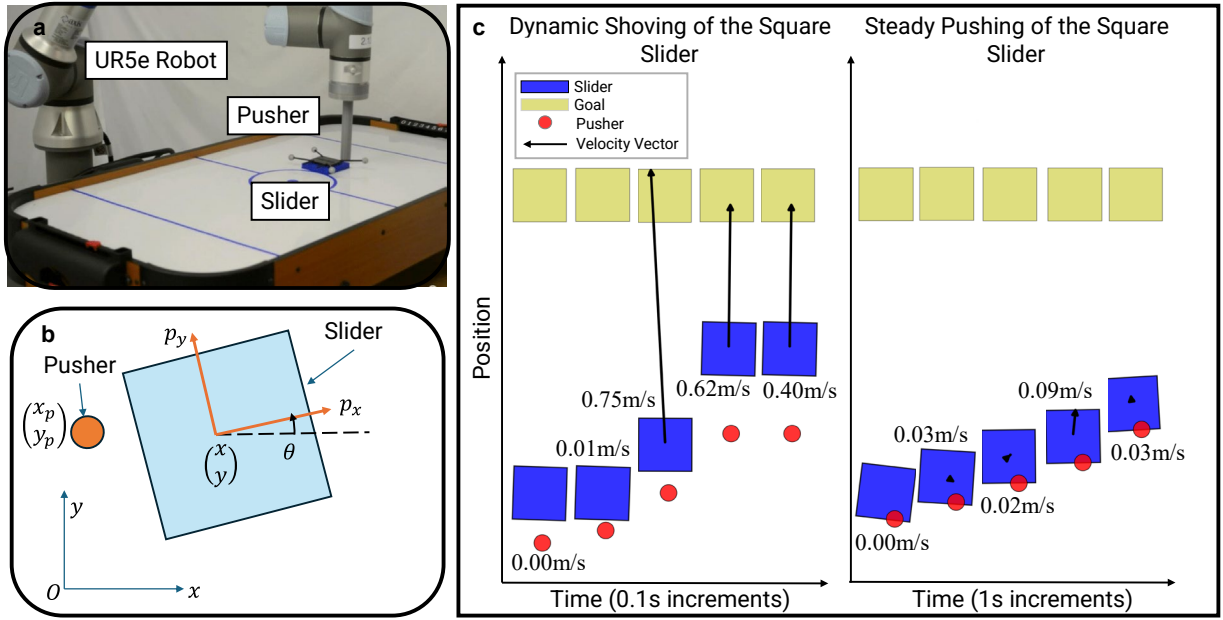

**Fig. 5 Planar pushing with a square slider**
(**a**) shows a photograph of the hardware setup. (**b**) shows a schematic of the planar pushing system with a square slider and the definition of two coordinate frames: a global frame, $O$, and a local frame. (**c**) shows the results from two experiments with the same slider goal positions. The left plot showcases shoving, while the right plot shows a standard pushing behavior. See Supplementary Movie 4 for a visualization.

Assuming high-fidelity velocity control, we can represent the actuator dynamics as two integrators in the task space where inputs appear linearly. In Laplace form,

$$x_p = \frac{1}{s} u_x, \qquad y_p = \frac{1}{s} u_y \qquad (13)$$

Page **12** of **31**

where the inputs to the robotic pusher are desired velocity commands: $u_x = \dot{x}_{pd}, u_y = \dot{y}_{pd}$.

Like the rimless wheel, the use of a viscoelastic contact model allows us to satisfy the conditions required for CCK formulation. In this case, the plant is the slider with states $\mathbf{q} = [x, y, \theta, \dot{x}, \dot{y}, \dot{\theta}]^T$. The dynamics of this plant are not directly affected by control inputs $\mathbf{u} = [u_x, u_y]^T$ which instead lead to changes in the pusher's state only, $\mathbf{p} = [x_p, y_p]^T$. It is only through the pusher's position that the plant is affected, specifically when the pusher penetrates the slider and induces a contact force.

Leveraging a Deep Koopman Network (DKN)[42,43] for constructing embedding observables, CCK-MPC is applied to the planar pushing system. The resulting controller can produce dynamic solutions to the standard planar pushing task while tracking a goal trajectory for the slider. In addition to the slow and steady pushing that has been reported in prior work[11,44–47], our controller can dynamically shove the slider to move it quickly towards a goal. This behavior emerges when it is favorable to rapidly reduce the system's state error, at the expense of a more aggressive control input. Such behavior can be created by varying the weights penalizing state tracking error vs. control cost in the MPC cost functional, as well as by varying the reference trajectory being targeted by the controller.

This CCK-MPC control has been implemented on the robot system shown in Fig. 5. Penalizing control efforts more than state-tracking errors encourages CCK-MPC to perform a steady pushing behavior, similar to prior work (Fig. 5c). The opposite case - where control is cheap and state errors are costly - encourages the controller to move the slider towards the goal as quickly as possible, via a shoving behavior (Fig. 5d). These two behaviors were generated based on the same CCK-MPC controller: both setups have the same goal but have different weight values in their control functions. In the case of shoving, the pusher moves rapidly towards the slider. This accelerates the slider towards the goal such that contact is broken, allowing for a lower tracking error. This contrasts with a steady push, in which the pusher maintains consistent contact with the slider and moves at a slower speed, taking a longer time to catch up. These results indicate that CCK-MPC can make a strategic decision, continuous pushing or intermittent shoving, depending on the performance goal (Supplementary Movie 4).

CCK-MPC can also discover control sequences that make-and-break contact with a circular slider (Fig. 6). Unlike the prior work where a sequence of contact changes, or guidance to the pusher, was provided off-line, CCK-MPC can find it on its own; it can determine when and where the pusher should contact the slider and generates a path that repositions the pusher in an optimal manner. In Fig.6c, we initialized a horizontal pushing task with the pusher located on the wrong side of the slider. The robot can successfully reposition the pusher from an unfavorable initial configuration and move the slider towards the goal. Supplementary Movie 5 includes this experimental result, as well as an example of the controller recovering from external disturbances while accomplishing the tracking task.

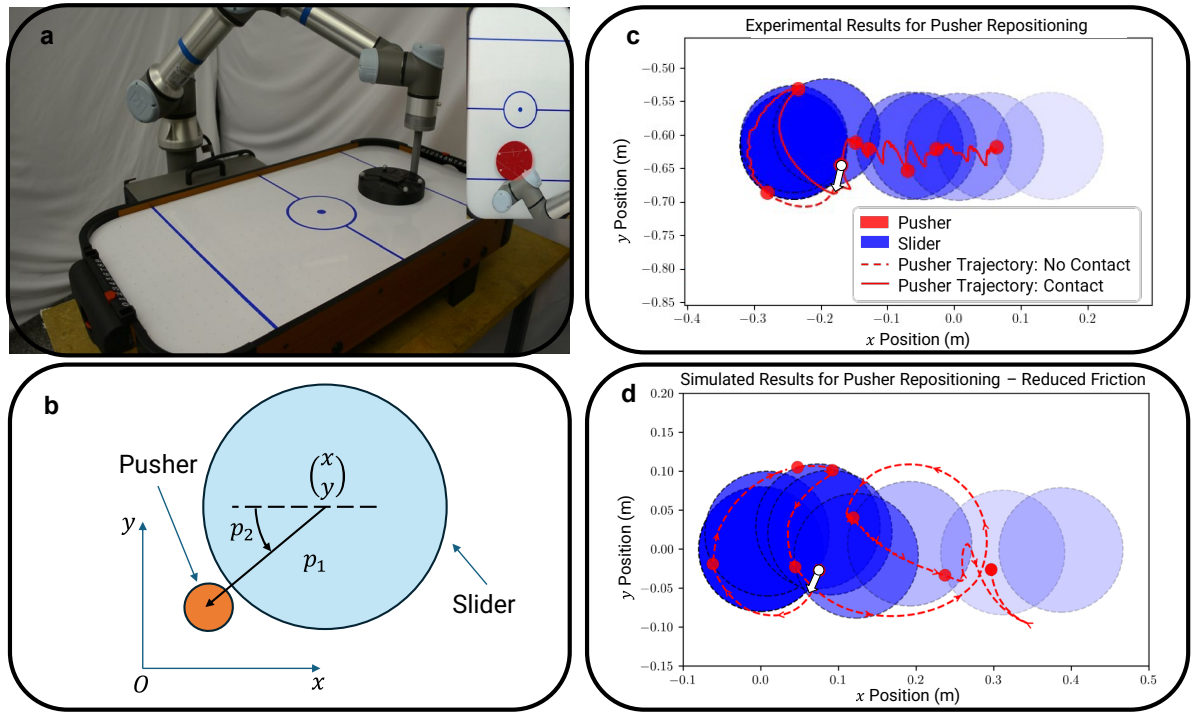

**Fig. 6 Planar pushing with a circular slider**
(**a**) shows a photograph of the experimental hardware setup. (**b**) shows a schematic of the planar pushing system with a circular slider. Two reference frames are presented: a global frame, $O$, and a local polar frame. (**c**) presents results from a hardware experiment in which the goal is to move the slider to the right. This goal is made challenging by initializing the pusher such that it is to the right of the slider, indicated by a white circle, meaning that it needs to reposition itself to complete the task. (**d**) presents the same task, except accomplished in simulation where the true friction is deliberately set to be lower than that used to generate the Koopman. Note that both (**c**) and (**d**) have been stretched in the $x$ direction to allow for the movement of the slider to be more easily seen.

To explore the performance of CCK-MPC in the presence of modeling errors, the same rightward-pushing task was implemented in simulation with a significantly reduced value of floor friction (Fig. 6d). Due to the lower-than-modeled friction, the slider tends to overshoot, and the duration of contact between the slider and pusher is brief and intermittent, leading to dribbling. Despite the modeling error in the friction parameter, CCK-MPC can recover and reasonably track the goal (Supplementary Movie 6).

Finally, to verify that the controller can run at real-time rates, we recorded the runtime for each iteration on a consumer laptop with an Intel i7-11800H. Gurobi[48] was used to solve the CCK-MPC optimization problem, and the code was written in Python. Figure 7 shows the distribution of runtimes for the control loop for a pair of simulated experiments: one with the square slider, and one with the circle slider. Both ran with mean runtimes of 5.2ms, which is sufficient for reactive, real-time operation.

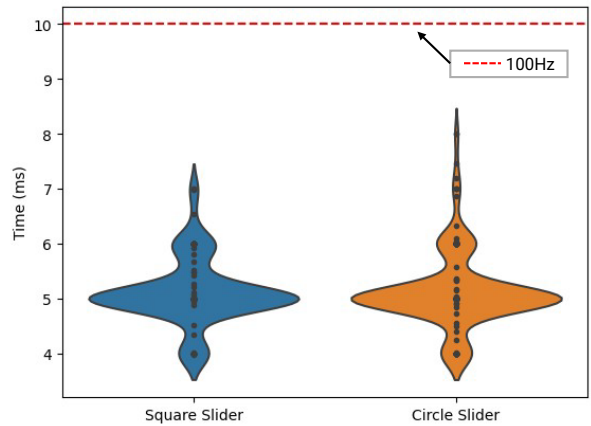

**Fig. 7 Distribution of CCK-MPC runtimes for manipulating square and circle sliders**
The distribution of individual control loop runtimes is shown for a pair of simulated experiments. 118 iterations are presented for the square slider, and 198 are shown for the circle slider. Although the controllers run at average rates of approximately 190Hz, there are occasional iterations that demand longer computations. Nevertheless, the controller empirically performs at rates faster than 100Hz with reasonable consistency.

## Discussion

The salient feature of the proposed CCK-MPC is global linearization that facilitates state prediction beyond the borders of segmented dynamics. Unlike local linearization, trajectory linearization, and DMDc approaches, the Koopman model is valid in a broader state space spanning local regions governed by diverse dynamics. The accuracy of local linearization, although high in the vicinity of a reference point, degrades rapidly as the state moves further from the reference. Particularly poor is the accuracy beyond a contact mode boundary; the model locally linearized within the current segment of the governing equations has no information about the dynamics beyond the boundary[5]. In contrast, the CCK model used for MPC can predict dynamic behaviors beyond the boundary. The CCK-MPC controller discovers a strategy for rolling the wheel by predicting and evaluating multi-step movements. The CCK-MPC controller of the dynamic pushing and shoving task finds the right location of the pusher for contacting the slider. Local linearization is not able to perform this type of prediction and control optimization. Other approaches based on trajectory linearization and DMDc, too, have shown poor performance, as compared in Fig. 4.

Koopman operators have already been applied to robotics and have had significant impacts upon active learning[49] and modeling of complex, nonlinear, distributed systems, such as soft robots[36,50,51]. In contrast, the current work uses the operator theory as a methodology for creating a unified, global model that subsumes segmented contact dynamics, thereby enabling robots to find elaborate control strategies in real-time. This is a new direction of applying the Koopman operator theory to robotics and related fields.

The use of Koopman operators for subsuming segmented dynamics and constructing convex MPC control systems was made possible a) by extending the original Koopman operator theory to non-autonomous systems with control and b) by building causal physical models without discontinuities at mode boundaries. Modeling compliance at the interactions between a robot and the environment is the key to accomplishing these; the floor compliance for the rimless wheel and the pusher rod compliance for dynamic pushing are critical modeling requirements. Resolving the causal conflict between two interacting bodies entails some form of compliance at the interface[41]. This compliance also allows actuators to possess independent state variables – satisfying Modeling Property 2 of the CCK formulation. Crucially, in rimless wheel modeling, no compliance is assumed in the powertrain. Instead, the independence of the actuator state **p** from the wheel state **q** is underpinned by the compliance of the floor. This differs from methods where the powertrain compliance is assumed for satisfying the CCK requirements[37]. In the dynamic pushing and shoving task, too, the powertrain is assumed to be rigid, with the pushing rod's compliance allowing for an independent actuator state.

In both cases, the applicability of the CCK formulation hinges on the contact compliance. This raises the question of whether a CCK model is valid for systems with a significantly high contact stiffness. We must investigate the effect of contact stiffness upon prediction accuracy of Koopman models and the resulting control performance of MPC. In general, prediction accuracy degrades as the contact stiffness increases beyond a certain level. This phenomenon pertains to a few issues, including the problem of stiff equations [52] As the contact stiffness becomes significantly higher, shorter time steps are required for computation. For MPC computation, this incurs an overall shorter prediction time horizon. In the current work, the effect of high contact stiffness upon MPC control performance can be investigated through numerical experiments. Supplementary Discussion 1.2 shows simulation results on the MPC trajectory cost against the stiffness of the pusher for the dynamic pushing and shoving tasks. There is an upper limit for the contact stiffness to assure high fidelity MPC control. More details are available in the doctoral dissertation of an author[53].

Contact compliance also impacts the control fidelity of actuators. If the contact stiffness is very low, it may incur slower dynamics and phase lag in controlling the actuator, because the control of the contact force requires a large movement of the actuator. Therefore, the contact stiffness must not be too low. There is an acceptable range for the contact stiffness that meets the two conflicting requirements: lower stiffness for prediction accuracy and higher stiffness for control fidelity. The simulation result in the Supplementary Discussion 1.2 shows the trade-off between the two requirements and a recommended range of the contact stiffness. The use of a desirable contact compliance allows the CCK-MPC to improve its performance. We can treat the contact impedance as a design parameter that we can select for improved control. The tip of each rimless wheel spoke can be covered with viscoelastic materials, and the pusher of the manipulation robot may be replaced by a rod of tuned viscoelasticity. These viscoelastic materials can not only absorb impacts and store energy but can also modulate the dynamics of the system. An optimal design may be considered for the robot-environment contact compliance and damping so that the overall control performance can be maximized.

CCK-MPC opens new possibilities for advancing robotics. Non-prehensile manipulation is no longer restricted to quasi-static and quasi-dynamic processes. We can deal with fully dynamic scenarios, a predetermined sequence of contact conditions is no longer required[20–23], and the robot can make-and-break contact freely to optimize its task performance. CCK-MPC can generate diverse modes of manipulation with a unified cost functional. Two types of actions - continuous pushing and intermittent shoving - are achieved simply by altering the weighting matrices **Q** and **R**. CCK-MPC finds an optimal strategy and selects the mode of action in real-time. Both the rimless wheel and dynamic pusher can perform advanced tasks with minimal guidance and without restrictive conditions.

The CCK-MPC method also outperforms other Koopman-based methods. DMDc and the methods based on local linearization cannot turn the rimless wheel, as shown in Fig.4. We compared a bilinear modeling approach to our CCK-MPC, by applying it to the MPC control of dynamic pushing and shoving (Supplementary Discussion 1.4). While bilinear Koopman MPC achieved high prediction accuracy, the bilinear MPC took more than 10 times longer than the proposed CCK-MPC. Bilinear MPC is not convex optimization, so it is difficult to apply to real-time, reactive control.

There are known limitations to Koopman operator-based modeling. Fundamentally, truncating an infinite-dimensional Koopman model to a finite-dimensional model results in an approximation of the original system's dynamics. This incurs imperfect predictions that can lead to compounding errors in future state predictions. We have investigated the effect of ablation for the dynamic pushing and shoving tasks, where Deep Neural Networks (DNN) [36,42,43,54,55] were used, as described in the Methods section. In general, as the number of DNN observables increases, the prediction accuracy improves at the cost of increased computational load (Supplementary Discussion 1.5). In addition, spectral analysis[26] provides an effective method for truncation, and physical system modeling provides insights into informative observables[56]. How all these potential benefits could be incorporated into a learning-based approach for generating observables remains the topic of future work.

## Methods

This section presents key formulas of CCK-MPC and procedures for model construction and control implementation for each of the rimless wheel and dynamic pushing/shoving applications. Further details needed for reproducing our results are described in the Supplementary Methods 2.1-2.3.

### CCK Model Derivation

The construction of a CCK model is detailed below, focusing on the time evolution of lifted state variables, which is LTI. The complete proposition of the CCK modeling is stated as:

For a non-autonomous system $\mathcal{S}_\mathcal{N}$ given by Equations (4) and (5), where actuator dynamics possess state variables $\mathbf{p}_t \in P \subset \mathbb{R}^m$ that are independent of plant state variables $\mathbf{q}_t \in X_q \subset \mathbb{R}^{n-m}$ and where actuator input $\mathbf{u}_t \in U \subset \mathbb{R}^r$ appears linearly in the time evolution of the actuator state, a Control-Coherent Koopman model exists and is given by Equation (11), where $\mathbf{B}$ is a constant control matrix and $\mathbf{A}$ is the Koopman operator of the associated autonomous system $\mathcal{S}_\mathcal{A}$, which is assumed to exist and is valid for $\forall \mathbf{q}_t \in X_q$ and $\forall \mathbf{p}_t \in P$, including all the actuator states that can be driven by an arbitrary input $\mathbf{u}_t$ in $U$. The sampling interval of the discrete-time system is assumed small.

∎

The time evolution of actuator state $\mathbf{p}_t$ has been given by Equation (10), where the state $\bar{\mathbf{p}}_{t+1}$ of $\mathcal{S}_\mathcal{A}$ is converted to $\mathbf{p}_{t+1}$ of $\mathcal{S}_\mathcal{N}$ by adding $\mathbf{B}_p\mathbf{u}_t$. The time evolution of the embedding observables $\mathbf{g}_t$ requires the conversion of $\bar{\mathbf{g}}_{t+1}(\bar{\mathbf{p}}_{t+1}, \mathbf{q}_{t+1})$ to $\mathbf{g}_{t+1}(\mathbf{p}_{t+1}, \mathbf{q}_{t+1})$. Considering a small time-interval,

$$\mathbf{g}_{t+1}(\mathbf{p}_{t+1}, \mathbf{q}_{t+1}) \cong \bar{\mathbf{g}}_{t+1}(\bar{\mathbf{p}}_{t+1}, \mathbf{q}_{t+1}) + \frac{\partial \bar{\mathbf{g}}_{t+1}(\bar{\mathbf{p}}_{t+1}, \mathbf{q}_{t+1})}{\partial \bar{\mathbf{p}}_{t+1}} \Delta \mathbf{p}_{t+1} \qquad (14)$$

where, from Equation (10),

$$\Delta \mathbf{p}_{t+1} = \mathbf{p}_{t+1} - \bar{\mathbf{p}}_{t+1} = \mathbf{B}_p \mathbf{u}_t \qquad (15)$$

The Jacobian $\mathbf{G} = \frac{\partial \bar{\mathbf{g}}_{t+1}(\bar{\mathbf{p}}_{t+1}, \mathbf{q}_{t+1})}{\partial \bar{\mathbf{p}}_{t+1}}$ can be computed from Equation (9). As the block matrix $\mathbf{A}_{pp}$ is non-singular, we obtain

$$\bar{\mathbf{g}}_{t+1} = \mathbf{A}_{gp} \mathbf{A}_{pp}^{-1} (\bar{\mathbf{p}}_{t+1} - \mathbf{A}_{pg} \mathbf{g}_t) + \mathbf{A}_{gg} \mathbf{g}_t \qquad (16)$$

Therefore,

$$\mathbf{G} = \mathbf{A}_{gp} \mathbf{A}_{pp}^{-1} \qquad (17)$$

Substituting Equations (15) and (17) into (14) and using (9) yield

$$\mathbf{g}_{t+1} = \mathbf{A}_{gp} \mathbf{p}_t + \mathbf{A}_{gg} \mathbf{g}_t + \mathbf{A}_{gp} \mathbf{A}_{pp}^{-1} \mathbf{B}_p \mathbf{u}_t \qquad (18)$$

By replacing $\mathbf{A}_{gp} \mathbf{A}_{pp}^{-1} \mathbf{B}_p$ by $\mathbf{B}_g$, which is a constant matrix, and combining Equation (10), the CCK model Equation (11) is obtained. All parameter matrices are constant. This concludes the proof of the CCK formulation.

The system matrices $\mathbf{A}$ and $\mathbf{B}$ of the CCK model can be obtained in various ways. The matrix $\mathbf{B}_p$ in $\mathbf{B}$ can be obtained from the actuator dynamics separately and prior to the identification of the $\mathbf{A}$ matrix. Standard methods can be applied to determining the $\mathbf{A}$ matrix of $\mathcal{S}_\mathcal{A}$ and the matrix $\mathbf{B}_g$ can be constructed from $\mathbf{A}$ and $\mathbf{B}_p$, as shown above. Alternatively, a more data-driven method can be applied based on the CCK model structure. Let $\{(\mathbf{z}_t, \mathbf{u}_t); t = 0, \cdots, N_D\}$ be data generated from multiple sets of segmented dynamic equations for diverse contact conditions. The matrices $\mathbf{A}$ and $\mathbf{B}_g$ can be obtained from

$$(\mathbf{A}, \mathbf{B}_g) = \underset{\mathbf{A}, \mathbf{B}_g}{\operatorname{argmin}} \sum_{t=0}^{N_D - 1} \left\| \mathbf{z}_{t+1} - \left( \mathbf{A} \mathbf{z}_t + \begin{bmatrix} \mathbf{B}_p \\ \mathbf{B}_g \end{bmatrix} \mathbf{u}_t \right) \right\|^2 \qquad (19)$$

where matrix $\mathbf{B}_p$ is predetermined.

The contribution of $\mathbf{B}_g$ is more significant as the time horizon of MPC extends further where the state equation (11) is computed repeatedly. For a short time horizon, on the other hand, the effect of $\mathbf{B}_g$ is negligibly small. See Supplementary Methods 2.1 and Supplementary Figure 5 for more details.

It should be noted that the standard DMDc does not utilize the CCK model structure where control input appears linearly in the original state equation. As a result, the control matrix **B** becomes state dependent. Approximating **B(x)** to a constant matrix is erroneous and misinforms the controller.

## MPC in Lifted Space

Given a CCK model like Equation (2) or (11) with embedding functions $\mathbf{g}(\mathbf{x}_t)$ for lifting the state to $\mathbf{z}_t(\mathbf{x}_t)$, a reference trajectory $\mathbf{x}_{t,\text{ref}}$ over a given time horizon of length $N$, and a measured current state $\mathbf{x}_{\text{measured}}$, an optimal control sequence is determined by solving the following linear MPC problem:

$$\min_{\mathbf{u}_0,\ldots,\mathbf{u}_{N-1}} \sum_{t=0}^{N-1} (\tilde{\mathbf{z}}_t^T \mathbf{Q} \tilde{\mathbf{z}}_t + \mathbf{u}_t^T \mathbf{R} \mathbf{u}_t) + \tilde{\mathbf{z}}_t^T \mathbf{Q} \tilde{\mathbf{z}}_t$$

$$\begin{aligned}
\text{s.t. } & \mathbf{z}_0 = \mathbf{z}(\mathbf{x}_{\text{measured}}) \\
& \mathbf{z}_{t+1} = \mathbf{A}\mathbf{z}_t + \mathbf{B}\mathbf{u}_t \\
& \tilde{\mathbf{z}}_t = \mathbf{z}_t - \mathbf{z}_{t,\text{ref}} \\
& \mathbf{z}_{t,\min} \leq \mathbf{z}_t \leq \mathbf{z}_{t,\max} \\
& \mathbf{u}_{t,\min} \leq \mathbf{u}_t \leq \mathbf{u}_{t,\max}
\end{aligned} \quad (20)$$

where weight matrices **Q** and **R** for each experiment and simulation are given in Supplementary Tables 2 and 4.

As the system is lifted, the number of state variables increases. Therefore, CCK lifted dynamical systems are often not controllable. However, MPC and optimal control do not require controllability. Therefore, the lack of controllability does not invalidate the use of Koopman models for MPC. One of the major benefits of CCK-MPC is that the optimal control computation is streamlined: convex optimization with no local minima and fast computation.

## Rimless Wheel Modeling

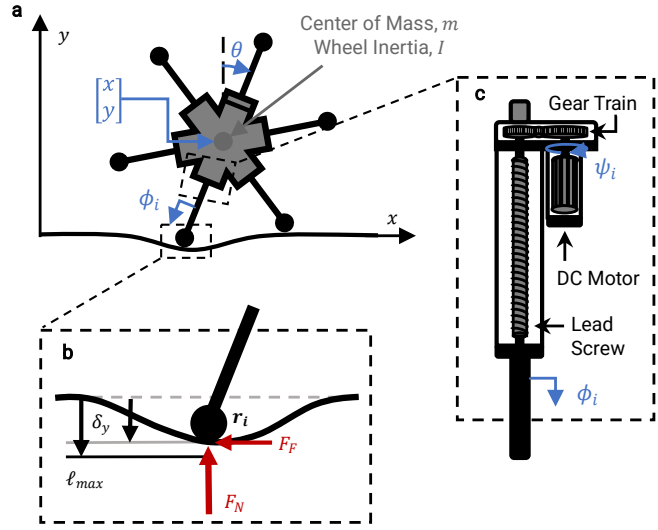

**Fig. 8 Schematic of rimless wheel contact model and actuator subsystem.**
(**a**) shows the rimless wheel's generalized coordinates and inertial elements associated with the hub. (**b**) illustrates key dimensions of the viscoelastic contact model. (**c**) shows the actuator components that contribute to the actuator inertia, as well as the relationship between the rotor angle $\psi_i$ and the spoke displacement $\phi_i$.

Figure 8 shows the schematic of the rimless wheel consisting of a hub and six spokes (legs) equally spaced at angle $\pi/3$. The configuration of the system is completely described with generalized coordinates, $\mathbf{x} = (x, y, \theta, \psi_1, \dots, \psi_6)^T$, where $\psi_i$ represents the angular displacement of the actuator driving the spoke of the $i$-th spoke sliding in the longitudinal direction. The inertial properties of the system are the central mass ($m$), the moment of inertia of the whole body ($I$), and the moment of inertia of each actuator rotor $I_m$. The kinetic energy is given by

$$T = \frac{1}{2}m(\dot{x}^2 + \dot{y}^2) + \frac{1}{2}I\dot{\theta}^2 + \frac{1}{2}I_m \sum_{i=1}^{6} \dot{\psi}_i^2 \qquad (21)$$

where the actuator rotor inertia $I_m$ includes the inertia of the gearing and the transmission mechanism reflected to the rotor axis. The spoke tip that translates in the radial direction is assumed massless. This implies that the whole-body inertia $I$ does not vary depending on the spoke lengths.

Potential energy is twofold. One is the gravity of the center mass $m$, and the other is the strain energy due to the viscoelastic interactions between the spokes and the floor. The former is represented with the potential energy $U = mgy$, while the latter is treated as part of generalized forces.

Figure 8b shows the viscoelastic contact model based on Khadiv et al.[38]. Let $k_y$ and $b_y$ be floor stiffness and damping, respectively, $\delta_y$ penetration depth, and $l_{max}$ its maximum. The normal reaction force is given by

$$F_N = -k_y \tan\left(\frac{\pi}{2l_{max}} \delta_y\right) - b_y |\delta_y|(\dot{\delta}_y) \tag{22}$$

Furthermore, let $\lambda$ be pseudo-Coulomb parameter and $\mu$ be coefficient of friction. Then the friction force is given by

$$F_F = -\frac{2}{\pi} \tan^{-1}\left(\frac{\dot{\delta}_x}{\lambda}\right) \mu F_N \tag{23}$$

The friction model is a smoothed Coulomb friction model whose magnitude depends on the normal force. When $\lambda$ approaches zero, the friction force becomes equal to that of the Coulumb model. Larger $\lambda$ values produce a smoother friction model and help decrease oscillatory friction forces. This contact model was used to recreate the ground reaction forces of a SURENA III bipedal robot[38].

Let $\mathbf{r}_i$ be the $x-y$ coordinates of the tip of the $i$-th spoke, which is a function of generalized coordinates $x, y, \theta, \psi_i$: $\mathbf{r}_i(x, y, \theta, \psi_i)$. No powertrain compliance is assumed, so that the spoke length $\phi_i$ is algebraically determined from $\psi_i$. The tip position relative to the floor is given by

$$\delta_{y,i} = \mathbf{r}_i \cdot \hat{e}_y - y_C \tag{24}$$

where $y_C$ is floor height, $\hat{e}_y$ is the unit vector along the $y$ axis, and $\delta_{y,i} < 0$ implies penetration into the floor. The tip velocities $\dot{\delta}_{x,i}, \dot{\delta}_{y,i}$, are obtained by differentiating the tip position $\mathbf{r}_i$. Substituting these into the floor-spoke interaction model yields the ground reaction force of the $i$-th spoke $\mathbf{F}_i = (F_{F,i}, F_{N,i})^T$.

The ground reaction forces are reflected to the generalized coordinates through the Jacobian:

$$\mathbf{Q}_i = \left(\frac{\partial \mathbf{r}_i}{\partial \mathbf{x}}\right)^T \cdot \mathbf{F}_i \tag{25}$$

The Lagrange's equations of motion are then given by

$$\mathbf{H}\ddot{\mathbf{x}} + \mathbf{G} = \sum_i \mathbf{Q}_i + \boldsymbol{\tau} \tag{26}$$

where $\mathbf{H}$ is the inertia matrix, $\mathbf{G}$ is the gravity of the center mass, and $\boldsymbol{\tau}$ is actuator torques, $\boldsymbol{\tau} = (0,0,0, u_1, \cdots, u_6)^T$.

The last six equations involved in the above equations of motion are actuator dynamics:

$$I_m \ddot{\psi}_i = \left(\frac{\partial \mathbf{r}_i}{\partial \psi_i}\right)^T \cdot \mathbf{F}_i + u_i \tag{27}$$

It should be noted that the actuator rotor torque $u_i$, that is the input to the actuator subsystem, appears linearly in the actuator dynamics, because the rotor inertia $I_m$ is constant. As the variables $\psi_i$ and $\dot\psi_i$ are independent state variables, this model satisfies the requirements for the CCK formulation.

Details about the computation of system matrix **A** and control matrix **B** as well as the construction of the observables used for the rimless wheel are described in Supplementary Method 1. Furthermore, details about the MPC formation and computation of the rimless wheel are in Supplementary Method 2.2.

Also, it should be noted that the CCK-MPC controller includes actuator dynamics. Unlike policies that fail to transfer from simulation to hardware due to neglected actuator dynamics[57], CCK-MPC generates control commands that are directly compatible with actuator-level control. The same applies to a low-level whole-body controller that bridges the gap between low dimensional trajectories or infrequent MPC computations and the motor commands.

## Dynamic Pushing/Shoving Modeling

The dynamics of the planar pusher system are governed by a) equations of motion of a slider, including b) a slider-floor friction model and c) a Koopman-compatible soft contact model between the slider and a pusher. The equations of motion are

$$m_s \begin{bmatrix} \ddot{x} \\ \ddot{y} \end{bmatrix} = (\mathbf{F}_{\text{fric}} + \mathbf{F}_{\text{push}}) \qquad (28)$$
$$I_s \ddot\theta = (M_{\text{fric}} + \mathbf{CP} \times \mathbf{F}_{\text{push}})$$

where $\ddot{x}, \ddot{y}, \ddot\theta$ are the accelerations of the slider, while $I_s$ and $m_s$ are the moment of inertia and mass of the slider, respectively. **CP** is the vector from the slider's center of mass to the pusher. $\mathbf{F}_{\text{push}}$ and $\mathbf{F}_{\text{fric}}$ are the forces due to contact with the pusher and friction with the ground, while $M_{\text{fric}}$ is the resultant torque from friction with the ground. The viscous friction model used in this example is such that:

$$\mathbf{F}_{\text{fric}} = -b_s^t \begin{bmatrix} \dot{x} \\ \dot{y} \end{bmatrix} \qquad (29)$$
$$M_{\text{fric}} = -b_s^r \dot\theta$$

The values for $b_s^t$ and $b_s^r$ was selected based on fitting to data on the physical system, but we note that the use of a viscous friction model is inherently an approximation to the underlying dynamics.

For representing $\mathbf{F}_{\text{push}}$, the distance between the pusher coordinates $x_p, y_p$ and the nearest edge of the slider must be obtained. Depending on the shape of the slider, we employ a different expression. For the square object, the pusher tip position is represented relative to the local coordinate system of the slider as $p_x, p_y$, as shown in Fig. 5a. Converting to these local coordinates, the force applied from the pusher to the slider, $\mathbf{F}_{\text{push}}$, is given by

$$\begin{bmatrix} p_x \\ p_y \end{bmatrix} = \begin{bmatrix} \cos\theta & \sin\theta \\ -\sin\theta & \cos\theta \end{bmatrix} \left( \begin{bmatrix} x_p \\ y_p \end{bmatrix} - \begin{bmatrix} x \\ y \end{bmatrix} \right) \qquad (30)$$

$$\mathbf{F}_{\text{push}} = \begin{cases} \mathbf{0} & , \Delta x_s \leq 0 \\ k_c \Delta x_s \begin{bmatrix} \cos\theta \\ \sin\theta \end{bmatrix} & , \Delta x_s > 0 \end{cases}$$

where $k_c$ is the contact stiffness, and $\Delta x_s = p_x + \frac{L}{2}$ with $L$ the length of the slider is the perpendicular distance that the pusher penetrates the slider. Stiffness $k_c$ was chosen after experimenting with a range of values in simulation.

For a circular slider, a polar coordinate system is used, and the position of the pusher is represented with respect to the distance from the center of the circular slider, $p_1$, and the polar coordinate angle $p_2$, as shown in Fig. 6a.

The dynamics of the pusher itself are more straightforward than those of the slider, thanks to the assumption of high gain velocity control. That is, the velocity of the pusher in the global frame is defined by the velocity commanded by the control input.

Thanks to the compliant contact dynamics, the physical model given in Equations (28-30) is Koopman-compatible. We can lift the planar pushing system with observables to obtain a globally linear model. For this manipulation system, we leverage a Deep Koopman Network (DKN) in order to learn observables $\mathbf{g}$ for lifting linearization[42,43]. With the resulting Koopman model, we can apply linear MPC to track a reference trajectory $(x, y)_{\text{ref}}$. The cost functional consists of the tracking error penalty with weight $\mathbf{Q}$ and the input cost penalty with weight $\mathbf{R}$, as shown in Equation (20).

We generated training data for the construction of effective Koopman observables from the nonlinear dynamics described in Equations (28-30). The network structure used in this work is based upon that by Lusch et al.[42], and the loss function consists of the sum of the squared errors associated to the independent state $\mathcal{L}_{\text{state}}$ and that of the observables $\mathcal{L}_{\text{obs}}$.

$$\mathcal{L} = \alpha_{\text{obs}} \mathcal{L}_{\text{obs}} + \alpha_{\text{state}} \mathcal{L}_{\text{state}} \tag{31}$$

where $\alpha_{\text{obs}}$ and $\alpha_{\text{state}}$ are weights. Specific hyperparameter values can be found in Supplementary Table 3.

The use-case of convex MPC requires us to pass the original state vector through the lifting process as a subset of the observables. In this way, our work differs from that presented by Lusch et al.[42]. We do not require a loss associated with the reconstruction of the state variables, since the state is explicitly included in the lifted representation. We also found that we did not need to include any regularization terms.

The state inputs used while training our DKN does not consider the slider's position: $x, y$. This is because the change in the slider's velocity is independent of its current position in the world: the position is not needed while calculating the dynamics. We therefore reduce the dimension of the DKN's input by two. Predictions of $x$ and $y$ are still needed to evaluate the cost function in CCK-MPC, and this is done through a trapezoidal integration scheme based on the velocity predictions from the DKN.

### Dynamic Pushing/Shoving Implementation

The full pseudocode algorithm for the proposed controller can be seen in Supplementary Fig. 6. The system's current state is first lifted to its higher dimensional representation, $\mathbf{z}_t$. While this is a nonlinear operation, note that it is being done outside of the optimization loop within CCK-MPC, as shown in Equation (20).

The state error cost matrix $\mathbf{Q}$ applies to errors in both independent state $x$ and observables $\mathbf{g}(\mathbf{x})$. However, any costs from the observable states are ignored by setting 0 to all the components of $\mathbf{Q}$ associated to the observables. This is a design decision made to ensure that the controller is focused on minimizing the errors in the system's true states, rather than in the lifted representation. Beyond this general approach, specific considerations had to be made for both the square and circle slider implementations. The values of $\mathbf{Q}$, $\mathbf{R}$ and the time horizon $N$ for each of the demonstrations in this paper can be found in Supplementary Table 4. Some additional comments regarding the practical implementation of CCK-MPC for the planar pushing task can also be found in Supplementary Methods 2.3.

Our CCK-MPC implementation realizes real-time computation of MPC for the planar pushing task, without any prior guidance to the pusher. No reference motion to guide the pusher is provided, no is any contact sequence defined in advance. The results in this paper demonstrate that a straightforward implementation of CCK-MPC, coded in Python, can perform at a sufficient rate for practical real-time usage. Our controller can empirically run in under 10ms, which corresponds to a potential rate of 100Hz, although it is worth noting that execution is not guaranteed to occur at these rates. In practice, our experiments were restricted to a control loop of 10Hz to align with the discretization timestep $\Delta t = 0.1$s used during the creation of the Koopman model. Running the controller at a faster rate – potentially up to its full 100Hz capability – is possible even with this discretization timestep but was found to not be necessary to demonstrate the desired performance.

In future work, additional speed-ups to enable frequencies beyond 100Hz may be possible by precomputing certain matrix operations offline[29]. It may also be possible to enforce guarantees on the time needed to compute the optimization, providing a useful time certificate for real-time use cases[58].

### Data Availability

The data obtained through simulation and hardware experiments are available on Zenodo with the identifier doi:10.5281/zenodo.17460547.

Specifically, the following material is available for review:
- Simulation data and control commands from the animation in Supplementary Movie 1.
- Simulation data, control commands, and MPC predictions in Fig. 3
- Simulation data and control commands from the animations in Supplementary Movie 2.
- Simulation data and control commands from the aggregated results in Fig. 4 and Supplementary Table 5, as well as the animations in Supplementary Movie 3.

- Raw state information collected during the dynamic shoving and pusher repositioning experiments demonstrated in Supplementary Movies 4 and 5, and figures 5C, 6D, and 6C, along with the control commands provided to the robot arm.

# Code Availability

Code for both systems have been made available on Github at https://github.com/jterrone1/Koopman-for-Rimless-Wheel and https://github.com/Cormac0/Koopman-for-Dynamic-Planar-Manipulation. Snapshots of the code are also provided on Zenodo with the identifier doi:10.5281/zenodo.17460547. The code is released under the MIT license.

# References

1. Pang, T., Suh, H. J. T., Yang, L. & Tedrake, R. Global Planning for Contact-Rich Manipulation via Local Smoothing of Quasi-Dynamic Contact Models. *IEEE Transactions on Robotics* **39**, 4691–4711 (2023).

2. Federico, S., Costanzo, M., De Simone, M. & Natale, C. Nonlinear Model Predictive Control for Robotic Pushing of Planar Objects With Generic Shape. *IEEE Robotics and Automation Letters* **10**, 3006–3013 (2025).

3. Hogan, F. R. & Rodriguez, A. Feedback Control of the Pusher-Slider System: A Story of Hybrid and Underactuated Contact Dynamics. in *Algorithmic Foundations of Robotics XII: Proceedings of the Twelfth Workshop on the Algorithmic Foundations of Robotics* (eds. Goldberg, K., Abbeel, P., Bekris, K. & Miller, L.) 800–815 (Springer International Publishing, Cham, 2020). doi:10.1007/978-3-030-43089-4_51.

4. Burke, J. V., Curtis, F. E., Lewis, A. S., Overton, M. L. & Simões, L. E. A. Gradient Sampling Methods for Nonsmooth Optimization. in *Numerical Nonsmooth Optimization: State of the Art Algorithms* (eds. Bagirov, A. M., Gaudioso, M., Karmitsa, N., Mäkelä, M. M. & Taheri, S.) 201–225 (Springer International Publishing, Cham, 2020). doi:10.1007/978-3-030-34910-3_6.

5. O'Neill, C. & Asada, H. H. Koopman Dynamic Modeling for Global and Unified Representations of Rigid Body Systems Making and Breaking Contact. in *2024 IEEE/RSJ International Conference on Intelligent Robots and Systems (IROS)* 709–716 (2024). doi:10.1109/IROS58592.2024.10801737.

6. Colgate, J. E. & Hogan, N. Robust control of dynamically interacting systems. *International Journal of Control* **48**, 65–88 (1988).

7. Koolen, T., De Boer, T., Rebula, J., Goswami, A. & Pratt, J. Capturability-based analysis and control of legged locomotion, Part 1: Theory and application to three simple gait models. *The International Journal of Robotics Research* **31**, 1094–1113 (2012).

8. Zhu, Y., Pan, Z. & Hauser, K. Contact-Implicit Trajectory Optimization With Learned Deformable Contacts Using Bilevel Optimization. in *2021 IEEE International Conference on Robotics and Automation (ICRA)* 9921–9927 (2021). doi:10.1109/ICRA48506.2021.9561521.

9. Kurtz, V. & Lin, H. Contact-Implicit Trajectory Optimization with Hydroelastic Contact and iLQR. in *2022 IEEE/RSJ International Conference on Intelligent Robots and Systems (IROS)* 8829–8834 (2022). doi:10.1109/IROS47612.2022.9981686.

10. Lynch, K. M. & Mason, M. T. Stable Pushing: Mechanics, Controllability, and Planning. *The International Journal of Robotics Research* **15**, 533–556 (1996).

11. Hogan, F. R., Grau, E. R. & Rodriguez, A. Reactive Planar Manipulation with Convex Hybrid MPC. in *2018 IEEE International Conference on Robotics and Automation (ICRA)* 247–253 (2018). doi:10.1109/ICRA.2018.8461175.

12. Graesdal, B. P. *et al.* Towards Tight Convex Relaxations for Contact-Rich Manipulation. *Robotics: Science and Systems 2024* https://doi.org/10.15607/RSS.2024.XX.132 (2024) doi:10.15607/RSS.2024.XX.132.

13. Wensing, P. M. & Orin, D. E. High-speed humanoid running through control with a 3D-SLIP model. in *2013 IEEE/RSJ International Conference on Intelligent Robots and Systems* 5134–5140 (2013). doi:10.1109/IROS.2013.6697099.

14. Dai, H., Valenzuela, A. & Tedrake, R. Whole-body motion planning with centroidal dynamics and full kinematics. in *2014 IEEE-RAS International Conference on Humanoid Robots* 295–302 (2014). doi:10.1109/HUMANOIDS.2014.7041375.

15. Wieber, P. Trajectory Free Linear Model Predictive Control for Stable Walking in the Presence of Strong Perturbations. in *2006 6th IEEE-RAS International Conference on Humanoid Robots* 137–142 (2006). doi:10.1109/ICHR.2006.321375.

16. Chen, Y.-M., Hu, J. & Posa, M. Beyond Inverted Pendulums: Task-Optimal Simple Models of Legged Locomotion. *IEEE Transactions on Robotics* **40**, 2582–2601 (2024).

17. Yamane, K. Systematic derivation of simplified dynamics for humanoid robots. in *2012 12th IEEE-RAS International Conference on Humanoid Robots (Humanoids 2012)* 28–35 (2012). doi:10.1109/HUMANOIDS.2012.6651495.

18. Wensing, P. M. *et al.* Optimization-Based Control for Dynamic Legged Robots. *IEEE Transactions on Robotics* **40**, 43–63 (2024).

19. Kuindersma, S. *et al.* Optimization-based locomotion planning, estimation, and control design for the atlas humanoid robot. *Auton Robot* **40**, 429–455 (2016).

20. Mason, M. T. Mechanics and Planning of Manipulator Pushing Operations. *The International Journal of Robotics Research* **5**, 53–71 (1986).

21. Zhou, J., Hou, Y. & Mason, M. T. Pushing revisited: Differential flatness, trajectory planning, and stabilization. *The International Journal of Robotics Research* **38**, 1477–1489 (2019).

22. Deits, R. & Tedrake, R. Footstep planning on uneven terrain with mixed-integer convex optimization. in *2014 IEEE-RAS International Conference on Humanoid Robots* 279–286 (2014). doi:10.1109/HUMANOIDS.2014.7041373.

23. Bledt, G. & Kim, S. Implementing Regularized Predictive Control for Simultaneous Real-Time Footstep and Ground Reaction Force Optimization. in *2019 IEEE/RSJ International Conference on Intelligent Robots and Systems (IROS)* 6316–6323 (2019). doi:10.1109/IROS40897.2019.8968031.

24. Reher, J. & Ames, A. D. Dynamic Walking: Toward Agile and Efficient Bipedal Robots. *Annu. Rev. Control Robot. Auton. Syst.* **4**, 535–572 (2021).

25. Aydinoglu, A., Wei, A., Huang, W.-C. & Posa, M. Consensus Complementarity Control for Multicontact MPC. *IEEE Transactions on Robotics* **40**, 3879–3896 (2024).

26. Mauroy, A., Mezic, I. & Susuki, Y. *The Koopman Operator in Systems and Control*. (Springer International Publishing, Cham, 2020). doi:10.1007/978-3-030-35713-9.

27. Koopman, B. O. Hamiltonian Systems and Transformation in Hilbert Space. *Proceedings of the National Academy of Sciences* **17**, 315–318 (1931).

28. Asada, H. H. Global, Unified Representation of Heterogenous Robot Dynamics Using Composition Operators: A Koopman Direct Encoding Method. *IEEE/ASME Transactions on Mechatronics* **28**, 2633–2644 (2023).

29. Korda, M. & Mezić, I. Linear predictors for nonlinear dynamical systems: Koopman operator meets model predictive control. *Automatica* **93**, 149–160 (2018).

30. Proctor, J. L., Brunton, S. L. & Kutz, J. N. Dynamic Mode Decomposition with Control. *SIAM J. Appl. Dyn. Syst.* **15**, 142–161 (2016).

31. Proctor, J. L., Brunton, S. L. & Kutz, J. N. Generalizing Koopman Theory to Allow for Inputs and Control. *SIAM J. Appl. Dyn. Syst.* **17**, 909–930 (2018).

32. Bruder, D., Fu, X. & Vasudevan, R. Advantages of Bilinear Koopman Realizations for the Modeling and Control of Systems With Unknown Dynamics. *IEEE Robotics and Automation Letters* **6**, 4369–4376 (2021).

33. Otto, S., Peitz, S. & Rowley, C. Learning Bilinear Models of Actuated Koopman Generators from Partially Observed Trajectories. *SIAM J. Appl. Dyn. Syst.* **23**, 885–923 (2024).

34. Folkestad, C. & Burdick, J. W. Koopman NMPC: Koopman-based Learning and Nonlinear Model Predictive Control of Control-affine Systems. in *2021 IEEE International Conference on Robotics and Automation (ICRA)* 7350–7356 (2021). doi:10.1109/ICRA48506.2021.9562002.

35. Iacob, L. C., Tóth, R. & Schoukens, M. Koopman form of nonlinear systems with inputs. *Automatica* **162**, 111525 (2024).

36. Feizi, N., Pedrosa, F. C., Jayender, J. & Patel, R. V. Deep Koopman Approach for Nonlinear Dynamics and Control of Tendon-Driven Continuum Robots. *IEEE Robotics and Automation Letters* **10**, 2926–2933 (2025).

37. Asada, H. H. & Solano-Castellanos, J. A. Control-Coherent Koopman Modeling: A Physical Modeling Approach. in *2024 IEEE 63rd Conference on Decision and Control (CDC)* 7314–7319 (2024). doi:10.1109/CDC56724.2024.10886771.

38. Khadiv, M. *et al.* Rigid vs compliant contact: an experimental study on biped walking. *Multibody Syst Dyn* **45**, 379–401 (2019).

39. Elandt, R., Drumwright, E., Sherman, M. & Ruina, A. A pressure field model for fast, robust approximation of net contact force and moment between nominally rigid objects. in *2019 IEEE/RSJ International Conference on Intelligent Robots and Systems (IROS)* 8238–8245 (2019). doi:10.1109/IROS40897.2019.8968548.

40. Geyer, H., Seyfarth, A. & Blickhan, R. Compliant leg behaviour explains basic dynamics of walking and running. *Proc. R. Soc. B.* **273**, 2861–2867 (2006).

41. Karnopp, D. C., Margolis, D. L. & Rosenberg, R. C. *System Dynamics: Modeling, Simulation, and Control of Mechatronic Systems*. (John Wiley & Sons, 2012).

42. Lusch, B., Kutz, J. N. & Brunton, S. L. Deep learning for universal linear embeddings of nonlinear dynamics. *Nature Communications 2018 9:1* **9**, 1–10 (2018).

43. Yeung, E., Kundu, S. & Hodas, N. Learning Deep Neural Network Representations for Koopman Operators of Nonlinear Dynamical Systems. in *2019 American Control Conference (ACC)* 4832–4839 (2019). doi:10.23919/ACC.2019.8815339.

44. Bauza, M., Hogan, F. R. & Rodriguez, A. A Data-Efficient Approach to Precise and Controlled Pushing. in *Proceedings of The 2nd Conference on Robot Learning* 336–345 (PMLR, 2018).

45. Moura, J., Stouraitis, T. & Vijayakumar, S. Non-prehensile Planar Manipulation via Trajectory Optimization with Complementarity Constraints. in *2022 International Conference on Robotics and Automation (ICRA)* 970–976 (2022). doi:10.1109/ICRA46639.2022.9811942.

46. Dengler, N., Ferrandis, J. D. A., Moura, J., Vijayakumar, S. & Bennewitz, M. Learning Goal-Directed Object Pushing in Cluttered Scenes with Location-Based Attention. Preprint at https://doi.org/10.48550/arXiv.2403.17667 (2025).

47. Del Aguila Ferrandis, J., Moura, J. & Vijayakumar, S. Nonprehensile Planar Manipulation through Reinforcement Learning with Multimodal Categorical Exploration. in *2023 IEEE/RSJ International Conference on Intelligent Robots and Systems (IROS)* 5606–5613 (2023). doi:10.1109/IROS55552.2023.10341629.

48. Gurobi Optiomization, LLC. Gurobi Optimizer Reference Manual. *Gurobi Optimization* https://www.gurobi.com/ (2024).

49. Abraham, I. & Murphey, T. D. Active Learning of Dynamics for Data-Driven Control Using Koopman Operators. *IEEE Transactions on Robotics* **35**, 1071–1083 (2019).

50. Bruder, D., Fu, X., Gillespie, R. B., Remy, C. D. & Vasudevan, R. Data-Driven Control of Soft Robots Using Koopman Operator Theory. *IEEE Transactions on Robotics* **37**, 948–961 (2021).

51. Haggerty, D. A. *et al.* Control of soft robots with inertial dynamics. *Science Robotics* **8**, eadd6864 (2023).

52. Burden, R. L., Faires, J. D. & Burden, A. M. *Numerical Analysis*. (Cengage Learning, 2015).

53. O'Neill, Cormac. Koopman Dynamic Modeling and Control for Robotic Systems Making and Breaking Contact. *Doctoral Dissertation in the Department of Mechnical Engineering* Massachusetts Institute of Technology (2025).

54. Han, Y., Hao, W. & Vaidya, U. Deep Learning of Koopman Representation for Control. in *2020 59th IEEE Conference on Decision and Control (CDC)* 1890–1895 (2020). doi:10.1109/CDC42340.2020.9304238.

55. Shi, H. & Meng, M. Q.-H. Deep Koopman Operator With Control for Nonlinear Systems. *IEEE Robotics and Automation Letters* **7**, 7700–7707 (2022).

56. Harry Asada, H. & Sotiropoulos, F. E. Dual Faceted Linearization of Nonlinear Dynamical Systems Based on Physical Modeling Theory. *Journal of Dynamic Systems, Measurement, and Control* **141**, (2018).

57. Neunert, M., Boaventura, T. & Buchli, J. Why off-the-shelf physics simulators fail in evaluating feedback controller performance - a case study for quadrupedal robots. in *Advances in Cooperative Robotics* 464–472 (WORLD SCIENTIFIC, 2016). doi:10.1142/9789813149137_0055.

58. Wu, L., Ganko, K. & Braatz, R. D. Time-certified Input-constrained NMPC via Koopman Operator. *IFAC-PapersOnLine* **58**, 335–340 (2024).

## Acknowledgements

Sumitomo Heavy Industries provided funds to support this work (CO, HA). This work was also supported by the National Science Foundation under Grant No. CMMI-2021625 (JT, HA) and the National Robotics Initiative Grant No. IIS-2133072 (JT, HA)

## Author Contributions

The CRediT taxonomy is used.

Conceptualization: CO, JT, HA
Methodology: CO, JT, HA
Investigation: CO, JT
Visualization: CO, JT
Funding acquisition: HA
Project administration: HA
Supervision: HA
Writing – original draft: CO, JT, HA
Writing – review & editing: CO, JT, HA

## Competing Interests

The authors declare no competing interests.